\documentclass[twocolumn, switch]{article} 
\usepackage{colortbl}
\usepackage{subcaption}
\usepackage{caption}
\usepackage{preprint}
\usepackage{graphicx}
\usepackage[utf8]{inputenc}
\usepackage[T1]{fontenc}
\usepackage{hyperref}
\hypersetup{breaklinks=true,colorlinks,bookmarks=false,linkcolor = purple,citecolor = orange,anchorcolor = black}
\usepackage[backend=biber,style=numeric,sortcites,sorting=nty,backref,natbib,hyperref]{biblatex}

\usepackage[autostyle=true]{csquotes}
\usepackage{amsmath, amsthm, amssymb, amsfonts}

\usepackage{doi}

\usepackage[utf8]{inputenc}	
\usepackage[T1]{fontenc}	
\usepackage{xcolor}		

\usepackage{booktabs} 		
\usepackage{nicefrac}		
\usepackage{microtype}		
\usepackage{lineno}		
\usepackage{float}			

\usepackage{lipsum}		
\usepackage{newfloat}
\DeclareFloatingEnvironment[name={Supplementary Figure}]{suppfigure}
\usepackage{sidecap}
\sidecaptionvpos{figure}{c}

\usepackage{titlesec}
\titlespacing\section{0pt}{12pt plus 3pt minus 3pt}{1pt plus 1pt minus 1pt}
\titlespacing\subsection{0pt}{10pt plus 3pt minus 3pt}{1pt plus 1pt minus 1pt}
\titlespacing\subsubsection{0pt}{8pt plus 3pt minus 3pt}{1pt plus 1pt minus 1pt}

\title{Evaluation of Human and Machine Face Detection using a Novel Distinctive Human Appearance Dataset}

\usepackage{xwatermark}
\newwatermark[firstpage,color=gray!90,angle=0,scale=0.28, xpos=0in,ypos=-5in]{*correspondence: \texttt{ngurkan@stevens.edu}}

\usepackage{authblk}

\author[1\thanks{\tt{ngurkan@stevens.edu}}]{Necdet Gurkan}
\author[2]{Jordan W. Suchow}

\affil[12]{School of Business, Stevens Institute of Technology}

\begin{document}

\twocolumn[ 
  \begin{@twocolumnfalse} 
  
\maketitle

\begin{abstract}
 Face detection is a long-standing challenge in the field of computer vision, with the ultimate goal being to accurately localize human faces in an unconstrained environment. There are significant technical hurdles in making these systems accurate due to confounding factors related to pose, image resolution, illumination, occlusion, and viewpoint \cite{merler2019diversity}. That being said, with recent developments in machine learning, face-detection systems have achieved extraordinary accuracy, largely built on data-driven deep-learning models \cite{wang2017detecting}. Though encouraging, a critical aspect that limits face-detection performance and social responsibility of deployed systems is the inherent diversity of human appearance. Every human appearance reflects something unique about a person, including their heritage, identity, experiences, and visible manifestations of self-expression. However, there are questions about how well face-detection systems perform when faced with varying face size and shape, skin color, body modification, and body ornamentation. Towards this goal, we collected the Distinctive Human Appearance dataset, an image set that represents appearances with low frequency and that tend to be undersampled in face datasets. Then, we evaluated current state-of-the-art face-detection models in their ability to detect faces in these images. The evaluation results show that face-detection algorithms do not generalize well to these diverse appearances. Evaluating and characterizing the state of current face-detection models will accelerate research and development towards creating fairer and more accurate face-detection systems.
\end{abstract}
\vspace{0.35cm}

  \end{@twocolumnfalse} 
] 



\section{Introduction}
Face-detection systems are becoming increasingly important as a first step of processing in many complex computer vision systems. Therefore, face detection is one of the most studied problems in the computer vision community. The execution of various face-based applications, from face recognition \cite{6909616,Schroff_2015} to photo-realistic face generation \cite{Bao_2017,Karras_2019,thomas2018persuasive}, social robot \cite{gordon2016affective}, face alignment \cite{6618919,6909614} and many other complex visual tasks depend on successful face detection. The purpose of a face detection is to determine whether there are any faces in an image and, if there are, to determine the location of each face. 

Computer vision systems trained using machine learning methods are widely used to support and make decisions in industry, healthcare, and government \cite{roussi_2020,muhammad2017facial}. Though the performance of such systems is often excellent \cite{raghavan2019mitigating}, accuracy is not guaranteed and must be assessed through careful measurement \cite{balakrishnan2020causal, grother2018ongoing}. The computational power of face-detection models comes from data-driven deep learning models that are trained using large datasets \cite{krkkinen2019fairface}. 

The difficulty in training face-detection algorithms is that the training data must provide sufficient balance and coverage to learn to represent human appearance \cite{yang2020towards}, and the images in a dataset must reflect the true diversity that exists in the world. Researchers have reported misbehavior of computer vision systems, including misidentification of minorities and other demeaning predictions \cite{e545256c2a9347cfa6debd4678287dd1, pmlr-v81-buolamwini18a}. Identifying biases correlated with people's appearances is particularly important for decisions that may have an impact on individuals’ lives. The measurement and correction of algorithmic and dataset biases is possible \cite{cavazos2020accuracy} and may help institutions progress towards fairer, more accountable, and more transparent environments. We must ensure computer vision systems are both inclusive of a wide spectrum of human appearances and accurate.

Fairness in computer vision recently started to receive increasing interest from different segments of academia and industry \cite{Wang_2020} because of the widespread deployment of A.I. technologies in daily life. Many existing models display bias because A.I. training datasets are inadequate in size and diversity \cite{Tommasi_2015} or they mirror human biases \cite{Caliskan_2017}. It is difficult to collect large datasets that reflect every aspect of human variation. On the other hand, designing and training a model that is invariant to all possible sources of variability is also impractical. Thus, any face analysis model suffers from bias and cannot generalize on sources of variation that are not explicitly modeled or are underrepresented in the training set \cite{terhorst2021comprehensive}.

The computer vision community has curated large-scale face image datasets \cite{Cao_2018,Guo_2016,liu2015faceattributes,Rothe2016DeepEO}. However, these image sets are biased toward Caucasian faces, and other races are significantly underrepresented. A recent study showed that most existing large-scale face databases are biased towards “lighter skin” faces compared to “darker” faces \cite{merler2019diversity}. Various unwanted biases in image datasets can easily appear due to biased selection, capture, and negative sets \cite{Georgopoulos_2020}. Most public datasets have been collected by scraping popular online websites, and the images retrieved from them are more frequently white people. 

Skin color, age, and gender are not the only ways in which people vary. Every human appearance reflects something unique to a person, including aspects of our heritage, identity, experiences, and other visible manifestations of self-expression. Human appearance varies for many reasons, including deliberate modifications to appearance (e.g., tattoos), adornments (e.g., piercings), genetic disorders that lead to deformity, catastrophic accidents and injuries, outward signs of group membership or culture, and many other factors \cite{Anderson-Frye}. These variations in human appearances raise the important question of how well face image datasets adequately represent the diversity of human appearances.

To stimulate research that recognizes the full variability of human appearance in furtherance of reducing bias in AI systems, we propose that face-detection algorithms be evaluated and compared using an image set with distinctive human appearances. To that end, we created the Distinctive Human Appearances dataset (DHA). This image set includes photographs of individuals with a wide range of distinctive appearances, including for example those with genetic disorders, victims of catastrophic accident, followers of unique fashion trends, those with body modifications, and members of cultural groups with distinctive appearances. We expect the DHA dataset and proposed associated fairness measure to be a resource for assessing face-detection algorithms, with evaluation results helping both to prioritize future research and to help better define the range of human appearances relevant to applications of face-detection technologies.

Our contributions are:
\begin{itemize}
   \item We introduce the Distinctive Human Appearances (DHA) dataset, which contains 1,000 images and associated annotations of low-frequency human appearances that capture diverse and complex variability in human appearance. 
   \item We show that nine face-detection models (2 traditional models, 4 deep learning models, and 3 cloud-based face analysis software applications) known to be highly accurate on benchmark evaluations fail to generalize to the DHA.
   \item In contrast, we show that human evaluators show excellent performance, with the primary failure mode being disagreement on what defines the bounds of a face. 
\end{itemize}

\section{Related Work}

Recently, numerous concerns have been raised regarding the accuracy and fairness of facial analysis systems. Popular face datasets such as Labeled Faces in the Wild \cite{LFWTech}, VGGFace2 \cite{Cao_2018}, MS-Celeb-1M \cite{Guo_2016}, and IMDB-WIKI \cite{Rothe2016DeepEO} are known to be imbalanced in both gender and skin color \cite{merler2019diversity}. To mitigate dataset biases in facial analysis systems, a number of attempts have been made to create more balanced datasets that are annotated for ethnicity and possibly other features. These include Racial Faces in the Wild (RFW) \cite{wang2018racial}, Balanced Faces in the Wild (BFW) \cite{Robinson_2020}, FairFace \cite{krkkinen2019fairface}, DiveFace \cite{Morales_2020}, and Diversity in Faces (DiF) \cite{merler2019diversity}. DiF was created using annotations defined from facial coding schemas, which provides quantative measures related to the intrinsic characteristics of faces \cite{merler2019diversity}. Notably, these more balanced datasets were created by subsampling  larger datasets while considering skin-tone, age, craniofacial ratios, and gender. Even though these datasets are an important step towards reflecting more of the diversity of human appearances, using a fixed set of category based labels for human appearances does not guarantee fair models \cite{Albiero_2020}. 

\renewcommand{\arraystretch}{1.4}
\begin{table}
\centering
\resizebox{\linewidth}{!}{%
\begin{tabular}{|c|c|c|c|c|c|c|} 
\cline{4-7}
\multicolumn{1}{c}{}                             & \multicolumn{1}{c}{}                                            &                     & \multicolumn{2}{c|}{\textbf{Gender}} & \multicolumn{2}{c|}{\textbf{Skin Type/Color}}  \\ 
\hline
\textbf{Dataset}                                 & \textbf{Source}                                                 & \textbf{\# of faces} & \textbf{Male} & \textbf{Female}      & \textbf{Darker} & \textbf{Lighter}             \\ 
\hline
\textbf{CelebA \cite{liu2015faceattributes}}                                           & \begin{tabular}[c]{@{}c@{}}CelebFace,\\LFW\end{tabular}         & 200K                & 42.0\%        & 58.1\%               & 14.2\%          & 85.8\%                       \\ 
\hline
\textbf{LFW \cite{LFWTech}}                                              & \begin{tabular}[c]{@{}c@{}}Web\\(Newspapers)\end{tabular}       & 13K                 & 77.4\%        & 22.5\%               & 18.8\%          & 81.2\%                       \\ 
\hline
\textbf{UTKFace \cite{Zhang_2017}}                                          & \begin{tabular}[c]{@{}c@{}}MORPH, CACD,\\Web\end{tabular}       & 20K                 & 52.2\%        & 47.8\%               & 35.6\%          & 64.4\%                       \\ 
\hline
\textbf{IMDB-Face \cite{Wang_2018_ECCV}}                                        & \begin{tabular}[c]{@{}c@{}}MegaFace,\\MS-Celeb-1M\end{tabular}  & 1.7M                & 55.0\%        & 45.0\%               & 12.0\%          & 88.0\%                       \\ 
\hline
\begin{tabular}[c]{@{}c@{}}\textbf{PubFig \cite{PMID:21383395}}\\\end{tabular} & \begin{tabular}[c]{@{}c@{}}Celebrity\\(Web Search)\end{tabular} & 13K                 & 49.2\%        & 50.8\%               & 18.0\%          & 82.0\%                       \\ 
\hline
\textbf{PPB \cite{pmlr-v81-buolamwini18a}}                                              & \begin{tabular}[c]{@{}c@{}}Gov. Official\\Profiles\end{tabular} & 1K                  & 55.4\%        & 44.6\%               & 46.4\%          & 53.6\%                       \\
\hline
\end{tabular}}
\vspace*{5mm}

\caption{Statistics of gender and skin type/color for popular face image datasets.}
\label{table:1}
\end{table}

While datasets are created to capture the diversity of human appearances, the data collection process can be biased by human and systematic factors, leading to distribution differences among datasets and reality, as well as between two datasets \cite{Lohia_2019}. For example, approximately one of three images in LFW \cite{LFWTech} include faces of individuals who are over 60 years old \cite{han2014age}, while CELEB-A \cite{liu2015faceattributes} include mostly light skin faces. Also, most of the existing face datasets contain politicians and celebrities \cite{Guo_2016,liu2015faceattributes,Rothe2016DeepEO,PMID:21383395}. These selections of sub-populations are not sufficient to represent diverse facial characteristics of humans and be generalize to the population. For instance, celebrities offer distinctive craniofacial attributes because of existence of make-up or plastic surgery, and politicians may be dominantly white. To ease representation bias, some datasets are collected by web search using certain keywords \cite{Zhang_2017}. These web search results may return stereotypes of searched keywords, such as celebrities, rather than diverse individuals from the public \cite{mcduff2019characterizing}. 

Face-detection systems that are trained with restricted appearances of a specific group  may learn into particular face features specific to that dataset. This issue occurs as under-representation or over-representation of specific arrangements of faces in many of the publicly available datasets \cite{merler2019diversity}. Table 1 shows the distribution of gender and skin type/color, source, and the number of faces for each widely used face image datasets. Aside from the FDDB and WiderFace datasets, popular face-detection training and testing datasets do not provide gender and skin color annotation. LFW, CelebA, and PubFig are highly narrowed towards lighter skin type/color. A similar bias is present with gender when grouped roughly into male and female. Building and training a computer vision system that is not representative of all possible face variability may generate bias models.

Over the last two decades, the National Institute of Standards and Technology (NIST) has curated the Face Recognition Vendor Test (FRVT), which is a standardized evaluation for accuracy and bias of face recognition systems \cite{Li_2019}. FRVT provides and quantifies demographic diversity for current face-recognition algorithms. However, this test is not proposed for iterative and fast evaluation of novel approaches in computer vision during development. In 2006, various academic and commercial face analysis algorithms tested displayed algorithmic and dataset bias such as algorithms developed in East Asia performed better on Asian individuals than Westerns individuals \cite{10.1145/1870076.1870082}. The reverse effect was present as well, the algorithms originated by Western developers performed better on Western individuals. Also, Phillips et al. \cite{10.1145/1870076.1870082} suggested that this inconsistency was caused by the different racial distribution in the training sets.

\begin{figure*}
\centering
\begin{center}
\includegraphics[width=17.2cm, height=6.5cm]{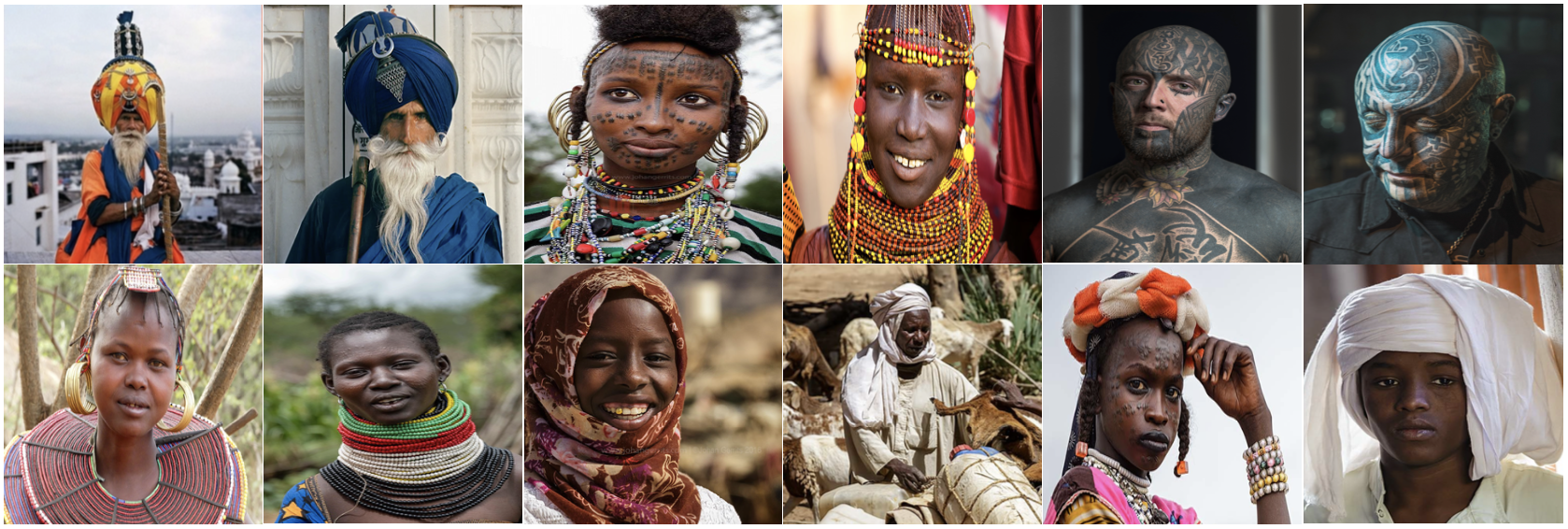}
\end{center}
   \caption{These images are samples from cultural and body modification sub-groups of the DHA and not randomly sampled from the DHA dataset. Photos used with permission from Johan Gerrits, Mark Hartman, Mark Leaver, and Rod Waddington.}
\label{fig:short}
\end{figure*}

In recent years, technology companies have begun providing on-demand cloud computing platforms and APIs to individuals, companies, and governments. Particularly, face recognition software is now built into most smart phones, airports, and drones. Face-detection algorithms are used by US-based law enforcement for surveillance and crime-prevention purposes \cite{gao}. \citeauthor{pmlr-v81-buolamwini18a} (\citeyear{pmlr-v81-buolamwini18a}) have shown systematic biases with the error rate being much higher for women than men in commercial face detection and tracking, face recognition, facial attribute detection, and facial expression services. Also, some commercial systems demonstrated algorithmic bias with regard to gender and skin color \cite{pmlr-v81-buolamwini18a}. In addition,  three commercial face-recognition systems that are evaluated in the NIST have been found to misidentify people of color, women and younger people, with many false positives \cite{10.1109/TIFS.2012.2214212}. These biases can easily cause problems in deployment, particularly in forensic settings. 

The research community has explored how faces may differ according to age, gender, and skin color, but these dimensions are insufficient to describe the whole distribution of human appearances. There are many other dimensions along which physical appearance is shaped, intentionally or unintentionally, such as tattooing and injury due to accidents. An overly simplistic view of human appearance may cause face-detection systems to function unfairly. Restricted diversity of face image datasets may have serious effects in computer vision applications. We believe this work will encourage academia and industry to develop fairer systems by considering the full distribution of diverse human appearances.

\section{The Distinctive Human Appearance Dataset}

We curated the Distinctive Human Appearances dataset to evaluate face-detection systems. This image set is designed to test the ability of face detection algorithms to localize human faces with diverse appearances. Therefore, the DHA represents less common appearances and those less likely to appear in extant image sets. Within this paper, we provide a novel dataset of human appearances that have received insufficient attention by the computer vision community, rather than focusing on appearances which have received sufficient attention. Collecting these images is not as simple as scraping Flickr or using any other methods to curate large datasets, unlike for example the Wider dataset.  It required manually searching the internet and various literatures to find factors that impact human appearances. The issue of capturing appearances with lower-frequency is not mentioned or widely considered in the computer vision community. The DHA dataset is a step toward curating a much larger dataset with a broad spectrum of human appearances.

The DHA dataset can be accessed at \url{https://sites.google.com/view/dha-dataset/} by submitting a form. We  provide a zip file that includes the images, bounding boxes and landmarks annotations, image source URLs and detailed descriptions of each sub-group of the DHA.

Images were collected by web search for terms related to appearance and group membership. Each image is cropped to contain exactly one face. All images are high resolution, with no occlusion and little variability in pose. These qualities are expected to result in good face-detection performance.

\cite{article,article123, yang2016wider}.


The DHA dataset comprises 1,000 still images in five broad categories that have a strong impact on human appearance. The dataset includes images of people who have experienced catastrophic accidents, suffer from genetic disorders, belong to any of several cultural groups, follow any of several fashion trends, or modified their body. The general distribution is 420 images from 25 different cultural groups, 267 images from 25 different genetic disorders, 135 images with body modifications, 120 images from three popular fashion trends, and 58 images from severe accidents. We provided detailed descriptions and information on these each sub-groups of the DHA in the zip file.  

We determined the sub-groups of the DHA by searching the internet for factors that influence human appearance significantly including severe accidents, diseases, body modifications, cultural groups, and interesting fashion trends. The size of sub-groups does not serve any particular purpose. Finding images for some of the sub-groups was easier than for others. For example, travel photography is a common way to capture images of people with different ethnicity and cultures, which helped us to include larger sub-group of such images. Conversely, finding images related to medical conditions is harder to discover because of privacy issues. Additionally, some of the medical conditions are rare, therefore, fewer examples by definition are available. This difficulty should not be confused with overlooking these individuals. We must ensure face detection algorithms are fair for everyone in our society, therefore, we need to make an effort to move towards a society that acknowledges individuals with lower-frequency appearances.

\begin{figure}[t]
\centering
    \includegraphics[width=5cm, height=4.6cm]{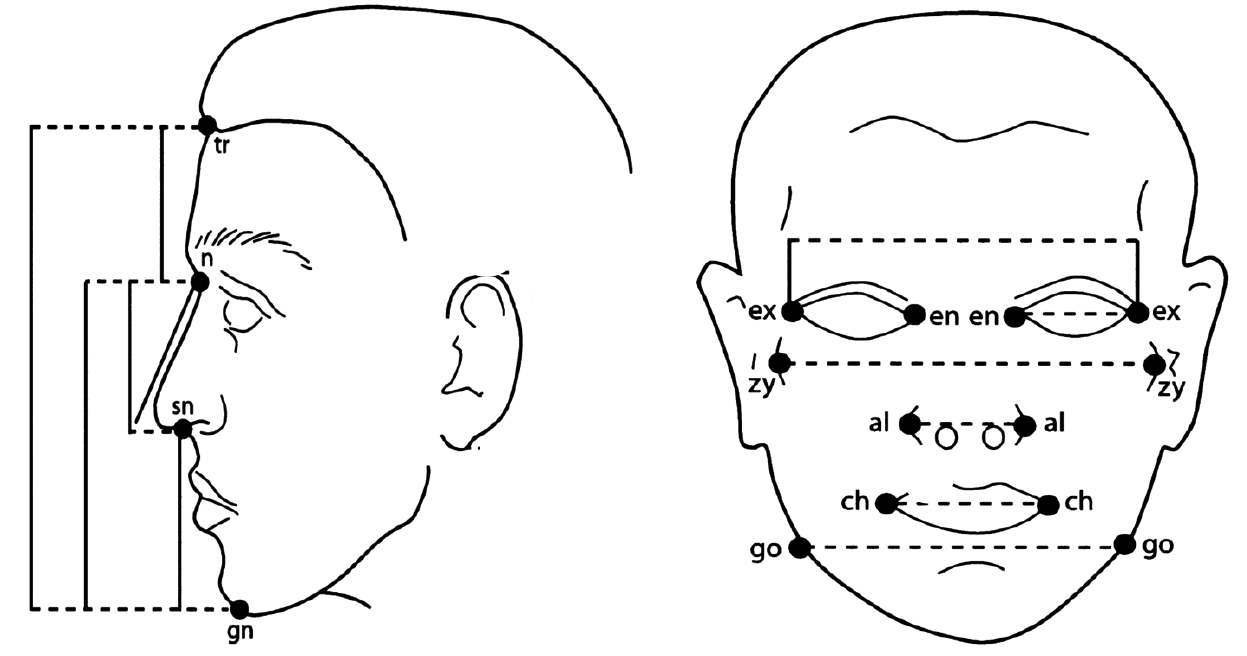}
    \caption{Measurements and landmarks on the lateral and frontal views of the face \cite{cranioface}.}
\end{figure}

\begin{table}[]
\begin{center}
\label{table:2}
\begin{tabular}{cccc}
\multicolumn{1}{l}{\textbf{Measurement}} & \multicolumn{1}{l}{\textbf{-1 SD}} & \multicolumn{1}{l}{\textbf{Mean}} & \multicolumn{1}{l}{\textbf{+1 SD}} \\ \hline
tr-n                                     & 15.2                               & 105.8                             & 196.4                              \\
tn-gn                                    & 48.2                               & 282.2                             & 516.3                              \\
n-gn                                     & 28.6                               & 176.9                             & 325.2                              \\
sn-gn                                    & 15                                 & 110.5                             & 206.1                              \\
zy-zy                                    & 29.4                               & 173.8                             & 318.3                              \\
go-go                                    & 23.8                               & 152.1                             & 280.5                              \\
en-en                                    & 8.4                                & 53.4                              & 98.5                               \\
en-ex                                    & 4.2                                & 47                                & 89.9                               \\
ex-ex                                    & 20                                 & 146.7                             & 273.4                              \\
n-sn                                     & 9.3                                & 67.6                              & 125.8                              \\
al-al                                    & 8                                  & 61.5                              & 115                                \\
ch-ch                                    & 6.5                                & 82.5                              & 158.5                             
\end{tabular}
\caption{Represents twelve craniofacial measures in pixels corresponding to different vertical and lateral distances in the face image.}
\end{center}

\end{table}

We collected bounding box annotations for each image using Amazon Mechanical Turk. We used a rectangular box to identify the location of a face. Each face was labeled by 10 annotators. We asked annotators to make their best judgments to place the bounding boxes that will include a face in an image. The smallest median value of face region bounding box size was $24\times21$. The largest median value of face region bounding box size was $1922 \times466$.  

Facial landmarks are standard quantitative references to craniofacial structure that differ across people \cite{cranioface}. Also, craniofacial structure can change from body modifications, injuries/accidents, surgical operations, etc. We used 12 anthropomorphic measurements to determine the morphological features of the craniofacial complex in the DHA dataset \cite{cranioface}. Since none of the face detection methods were sufficiently accurate to localize the facial landmarks in the DHA dataset, we manually annotated the facial landmarks shown in Fig. 2. 

The validation of current face-detection algorithms was performed on standard large datasets, where these algorithms are trained and validated on frequent human appearances. We show that these algorithms performed poorly and failed to generalize their execution to a novel unseen dataset, the DHA. However, face detection algorithms must generalize their performance to underrepresented appearances. For example, the DHA dataset can be used during validation and provide verification of their performance on unseen human appearances.

\begin{figure}[t]
\begin{subfigure}{8.3cm}
\includegraphics[width=8.3cm, height=7cm]{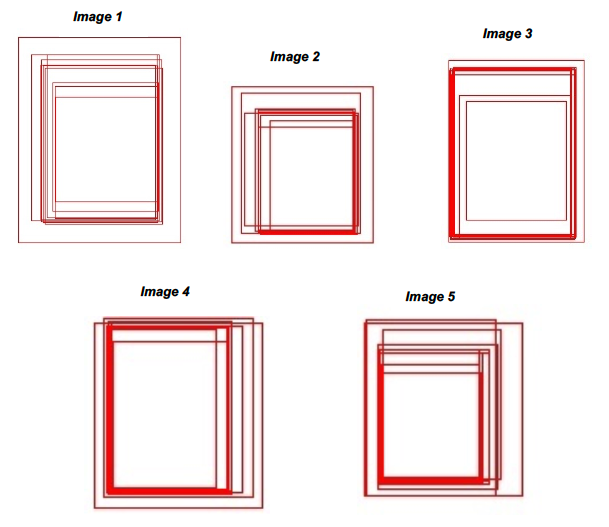}
\caption{Lowest variance}
\label{fig:subim1}
\end{subfigure}
\begin{subfigure}{8.3cm}
\includegraphics[width=8.3cm, height=7cm]{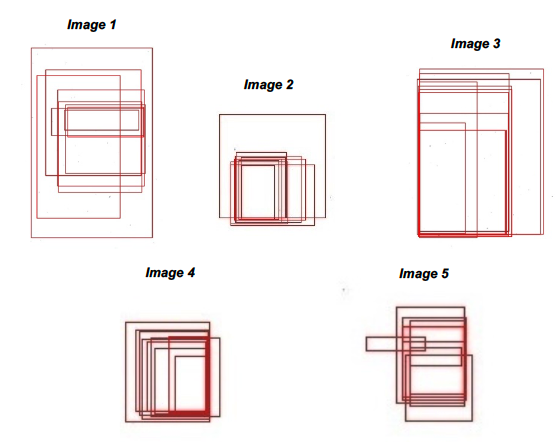}
\caption{Highest variance}
\label{fig:subim1}
\end{subfigure}
\caption{Examples of bounding boxes for 5 images with the highest and lowest variance performed by human annotators.}
\end{figure}




\section{Evaluating Human and Machine Detection on the DHA}
\textbf{Bounding box.} A bounding box is commonly considered correct if its Intersection Over Union (IoU) with a ground truth box is greater than 0.5 \cite{Everingham2009ThePV}. We consider the ground-truth as median values of each annotated side of bounding boxes. For face detection, an image is assigned a positive label if the IoU between an image and the ground truth bounding box is greater than 0.5; otherwise it is negative.

The shape of a human head can be approximated using two three-dimensional ellipsoids \cite{jain2010fddb}. Compared to other methods, such as evaluation of facial landmarks, orthographic projection of vertical ellipsoid, etc., the horizontal ellipsoid does not provide all available information about the features of the face region \cite{jain2010fddb}. In this paper, we keep the analysis as simple as possible to evaluate human annotation and machine performance on the DHA. The DHA dataset includes unique human appearances where it may even be considered difficult for humans to identify the face region accurately.

\subsection{Human Detection}  
In contrast to popular face datasets, the images in the DHA dataset have a distinctive distribution in clothing, skin colors and facial features. The diversity and uniqueness of DHA images may cause confusion for annotators when determining the bounds of a face during the task of annotation. For example, faces with several different body modifications can lead to different interpretations of where a bounding box should be placed. This is an important concern when human annotator performance forms the basis for judging the automated system's performance.

To obtain insights into these questions, we evaluate human annotation to validate human performance on the DHA dataset. Similar to evaluating machine detection, we assume the ground-truth as the annotated median values of each side of bounding boxes. Human annotators achieved 95\% accuracy with average IoU value of 0.82 and standard deviation of 0.15. The reason IoU rate is relatively imperfect is because of the ambiguity in the definition of a human face and bounding box. As shown in Fig. 3, human annotators displayed high variance on multiple DHA images. For example, the lowest IoU rate (\emph{M}=0.51, \emph{SD}=0.25) among the groups is generated by the sub-category of cultural images labelled “burka”, which is a concealing cover worn by some Muslim women. Some annotators assume that only the visible region of the face is the face or the expected face shape is the face. It can be an issue training face detection algorithms, particularly the algorithms that aims to generalize their high performance to the full distribution of human appearances. By demonstrating the issue of annotation, we aim to inform the computer vision community of annotation issues for future works.

\begin{figure}[t]
\centering
    \includegraphics[width=8.3cm, height=7cm]{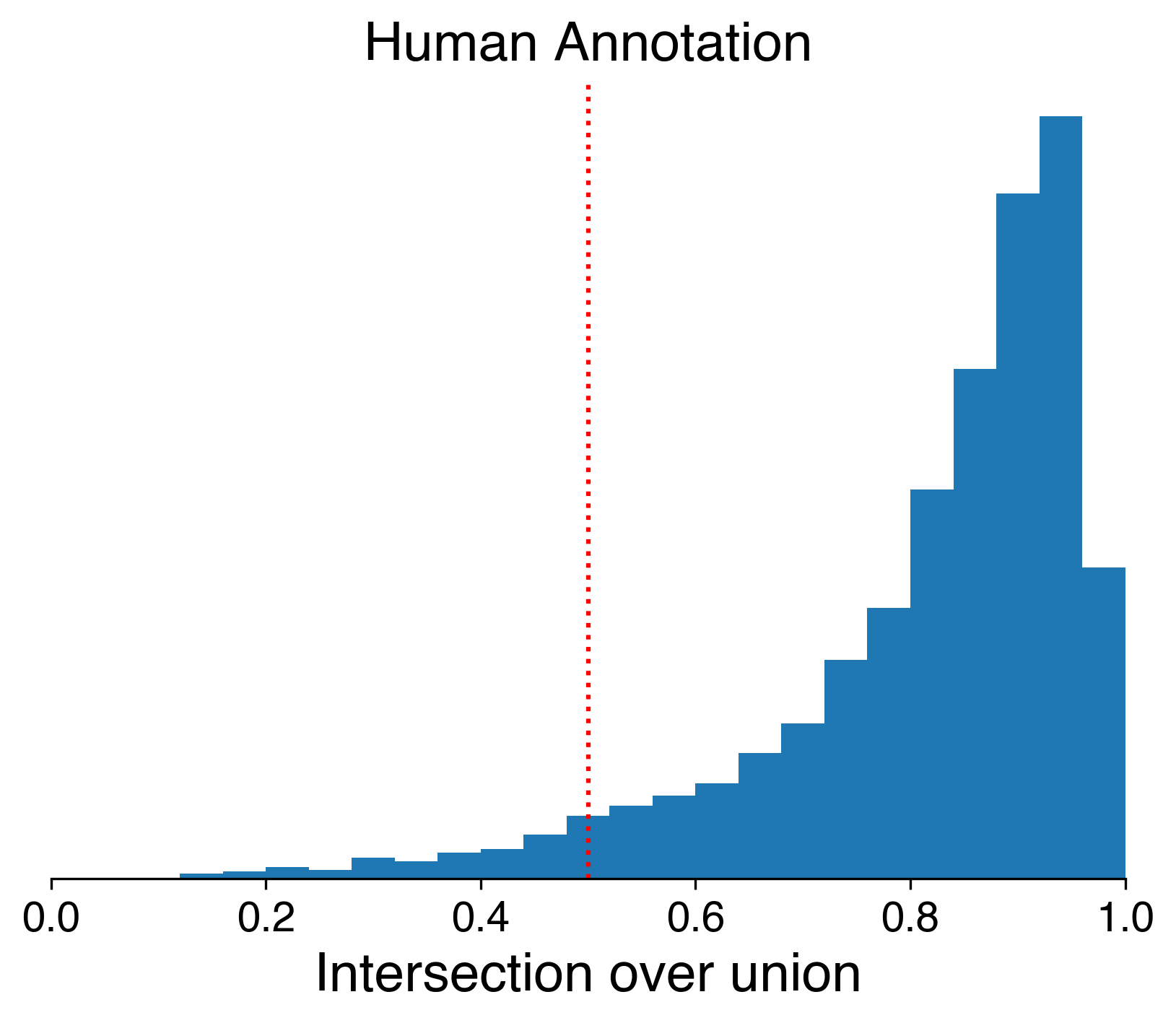}
    \caption{Histogram of each annotation. We consider median value of each side of bounding as a ground truth. Red dotted line represents threshold for each annotation to be accepted as positive or negative label.}
\end{figure}

\subsection{Machine Detection}  
We tested the following face-detection algorithms: Viola-Jones\cite{Viola04robustreal-time}, Dlib \cite{JMLR:v10:king09a}, Multi-Task Cascaded Convolutional Networks \cite{Zhang_2016}, RetinaFace \cite{RetinaFace},  {\texorpdfstring{S\textsuperscript{3}FD}{like this}}
 \cite{zhang2017s3fd}, 
Single Shot Multibox \cite{Liu_2016}, Google Cloud Vision \cite{google}, Amazon Rekognition \cite{aws}, and Microsoft Azure\cite{azure}. As previously mentioned, we used the ground truth as  median values of each annotated side of bounding boxes to evaluate selected face detection algorithms.  First, we evaluated the hand-crafted feature models. Viola-Jones\cite{Viola04robustreal-time} and Dlib \cite{JMLR:v10:king09a} detectors are widely used in many computer vision applications due to their fast processing speed and reasonably good accuracy \cite{Jabbar_2018,PMID:28268363,7477553,NEURIPS2018_d86ea612,Chung_2017}. Using Haar feature integrated model, the OpenCV implementation of Viola-Jones reached 34\% accuracy. Fig. 7 shows that the implementation of OpenCV library holds a high precision with low recall rate, which means that it has high accuracy in detected faces but lower recall. The second model, using HoG feature model, Dlib face detection achieved 65\% accuracy. Fig. 6 shows the precision-recall (PR) curves of the Dlib algorithm. Compared with the PR curve of the Viola-Jones algorithm shown in Fig. 7, the curve of Dlib is higher, which means algorithm has a higher precision than Viola-Jones when a higher sensitivity is required. The results show that these traditional methods are not robust in detecting faces in the DHA dataset compared to the results from other popular face datasets. 

\renewcommand{\arraystretch}{1.2}
\setlength{\tabcolsep}{4pt}
\begin{table}
\centering
\begin{tabular}{|lccc|} 
\hline
\rowcolor[rgb]{0.914,0.839,0.839} \textbf{Methods} & \begin{tabular}[c]{@{}>{\cellcolor[rgb]{0.914,0.839,0.839}}c@{}}\textbf{Detection }\\\textbf{ Rate }\end{tabular} & \begin{tabular}[c]{@{}>{\cellcolor[rgb]{0.914,0.839,0.839}}c@{}}\textbf{IoU Rate}\\\textbf{ Average }\end{tabular} & \begin{tabular}[c]{@{}>{\cellcolor[rgb]{0.914,0.839,0.839}}c@{}}\textbf{IoU Rate}\\\textbf{Standard }\\\textbf{ Deviation }\end{tabular}  \\
Human                                              & 95\%                                                                                                                       & 0.82                                                                                                               & 0.15                                                                                                                                      \\
Rekognition                                        & 95\%                                                                                                                       & 0.81                                                                                                               & 0.09                                                                       \\
\texorpdfstring{S\textsuperscript{3}FD}{like this}                                         & 91\%                                                                                                                       & 0.80                                                                                                               & 0.12                                                                                                                            \\
MTCNN                                              & 77\%                                                                                                                       & 0.75                                                                                                               & 0.17                                                                                                                                      \\
RetinaFace                                         & 65\%                                                                                                                       & 0.82                                                                                                               & 0.10                                                                                                                                      \\
Dlib                                               & 65\%                                                                                                                       & 0.68                                                                                                               & 0.10                                                                                                                                      \\
Google                                             & 58\%                                                                                                                       & 0.55                                                                                                               & 0.09                                                                                                                                      \\
Azure                                              & 57\%                                                                                                                       & 0.65                                                                                                               & 0.11                                                                                                                                      \\
MobilNet-SSD                                                & 49\%                                                                                                                       & 0.76                                                                                                               & 0.24                                                                                                                                      \\
Viola-Jones                                        & 34\%                                                                                                                       & 0.65                                                                                                               & 0.23                                                                                                                                      \\
\hline
\end{tabular}
\vspace*{5mm}
\caption{The comparison of face detection accuracy among the methods on the DHA dataset.}
\label{table:2}
\end{table}

For deep learning models, we select RetinaFace, {\texorpdfstring{S\textsuperscript{3}FD}{like this}}, MTCNN and MobilNet-SSD. These three deep learning models are widely used in many AI applications \cite{Deng_2019,hukkelaas2019deepprivacy, Liu_2017,autism,inproceedingsrobots,cheng2018surveillance,Ranjan_2019,Chung_2018, Zhou_2020_CVPR}. RetinaFace, {\texorpdfstring{S\textsuperscript{3}FD}{like this}}, MTCNN and MobilNet-SSD achieve excellent accuracy over the state-of-art techniques on the challenging FDDB and WIDER FACE benchmark for face detection and AFLW benchmark for face alignment \cite{Zhou_2020_CVPR, Zhang_2016, Liu_2016}.  However, RetinaFace achieved 65\% accuracy, MTCNN achieved 77\% accuracy, and MobilNet-SSD reached 49\% accuracy on our DHA dataset. {\texorpdfstring{S\textsuperscript{3}FD}{like this}} outperformed other state-of-art techniques and achieved 91\% accuracy. RetinaFace has a higher precision with a low recall, which means it can be used for tasks that require high accuracy but can tolerate lower recall. Compared with the PR curves of Viola-Jones, Dlib, MobilNet-SSD, MTCNN, and RetinaFace, {\texorpdfstring{S\textsuperscript{3}FD}{like this}} algorithm performs better than both traditional and deep learning algorithms at detecting the DHA faces.

\begin{figure}[t]
\begin{subfigure}{8.3cm}
\includegraphics[width=1\linewidth, height=5.5cm]{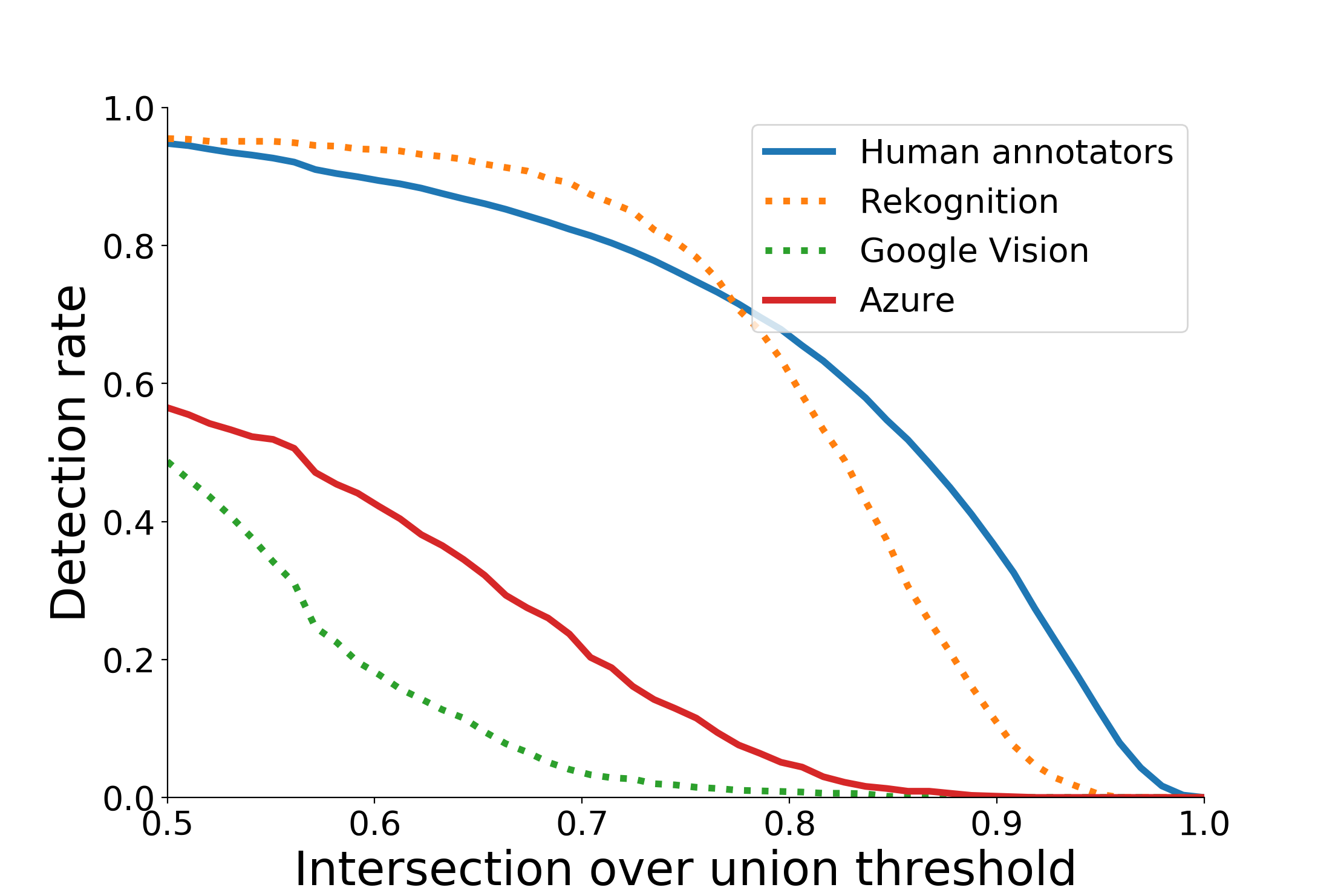} 
\caption{Cloud-based methods}
\label{fig:subim1}
\end{subfigure}
\begin{subfigure}{8.3cm}
\includegraphics[width=1\linewidth, height=5.5cm]{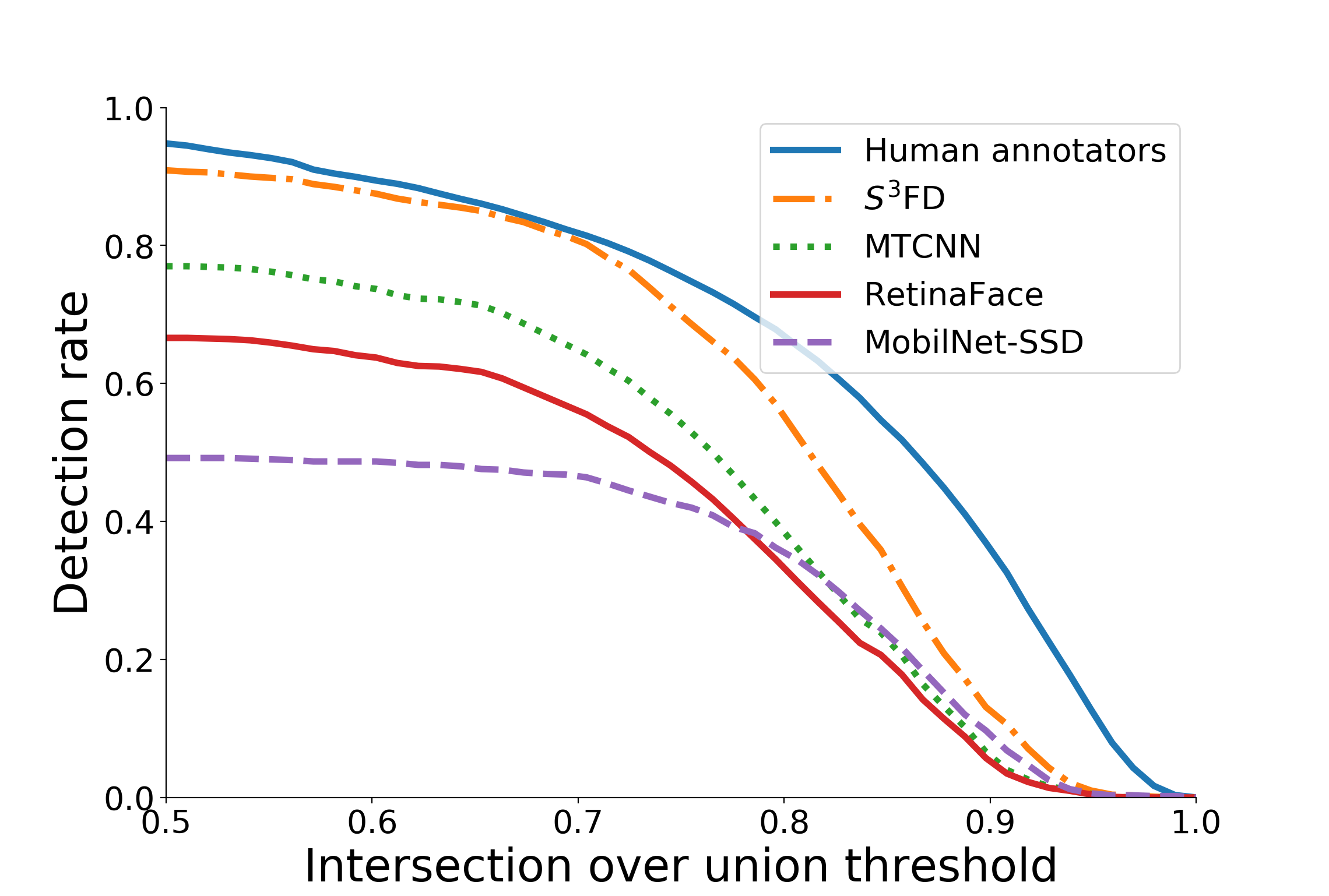} 
\caption{Deep learning methods}
\label{fig:subim2}
\end{subfigure}
\begin{subfigure}{8.3cm}
\includegraphics[width=1\linewidth, height=5.5cm]{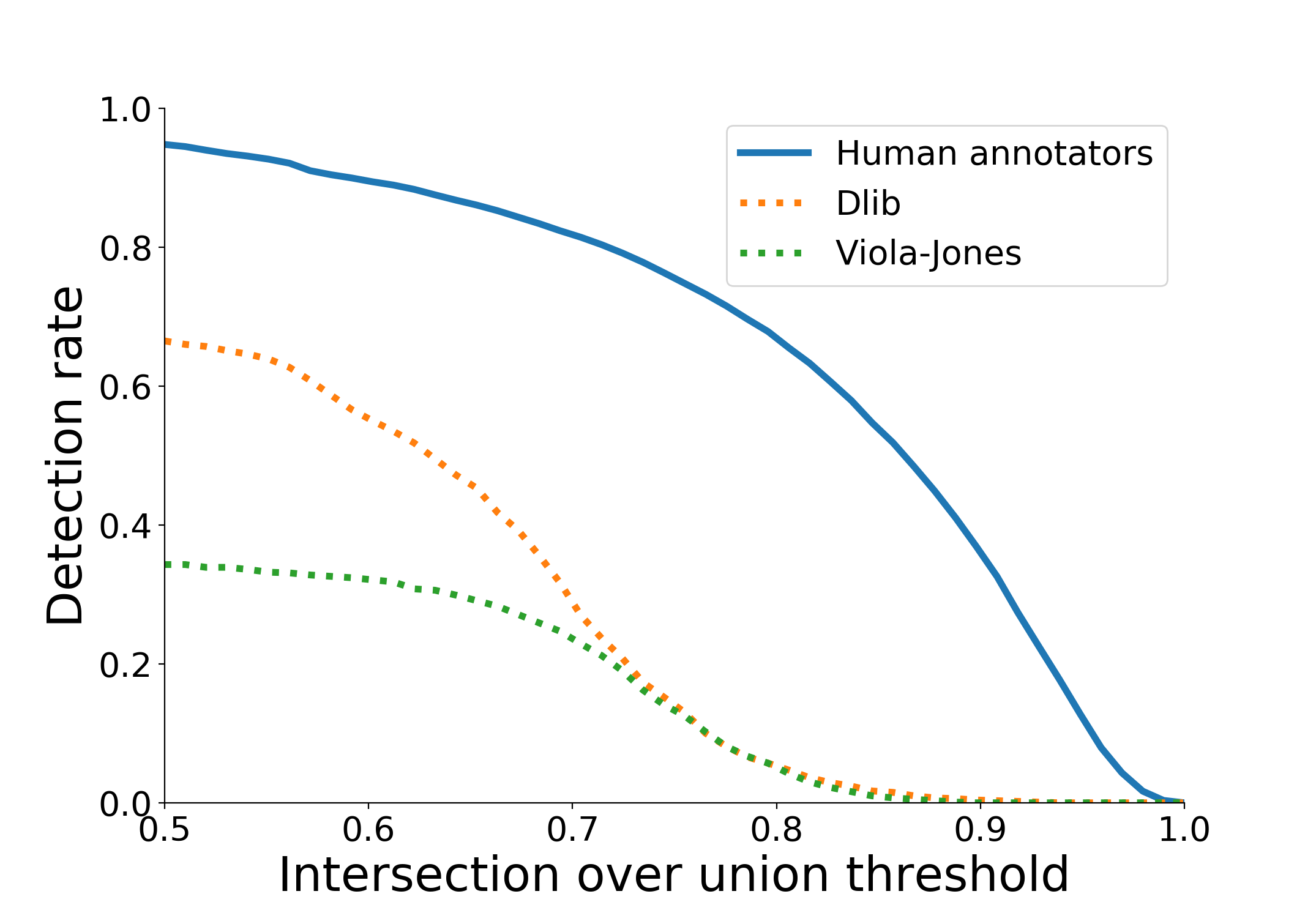}
\caption{Traditional methods}
\label{fig:subim3}
\end{subfigure}
\caption{Detection rate of face detection with the change of threshold value.}
\end{figure}

To evaluate Cloud-based face detection software implementations, we selected Amazon Rekognition \cite{aws}, Microsoft Azure \cite{azure}, and Google Cloud Vision \cite{google}. These cloud-based face detection software systems have been sold and used by security agencies, financial platforms, retail stores, and transportation systems \cite{microstory,googlestory,amazonstory}. The best performance is obtained by Amazon Rekognition with 95\% accuracy. Microsoft Azure and Google Cloud Vision performed similarly on the DHA dataset with accuracy rates of 57\% and 58\%, respectively. As shown in Table 3, the IoU rate for Google Cloud Vision is heavily weighted around 55\%, which is slightly above our ground truth. Microsoft Azure and Google Cloud Vision may perform worse if we slightly increase the ground truth box rate from 0.5. Compared with the PR curves of the Cloud-based face detection software, Amazon Rekognition provides long precision-recall curve that indicates an outstanding precision and recall rate.

Overall, Fig. 5 and Fig. 6 show that Amazon Rekognition performs best on our DHA dataset and demonstrates strong face detection capabilities even for tremendously diverse faces. {\texorpdfstring{S\textsuperscript{3}FD}{like this}} is ranked second but performed poorly compared to its performance on other datasets. RetinaFace, {\texorpdfstring{S\textsuperscript{3}FD}{like this}} and Amazon Rekognition have the highest precision rates with 98\%. On the other hand, Google Cloud Vision has the lowest precision rate with 74\%. Comparing the performance of detecting faces in DHA images, Amazon Rekognition has the highest recall rate with 97\%, and MobilNet-SSD has the lowest recall rate with 52\%. The results show that Amazon Rekognition is a superior implementation among the evaluated face detection implementations.  

\begin{figure}[t]
\centering
    \includegraphics[width=8.3cm, height=7cm]{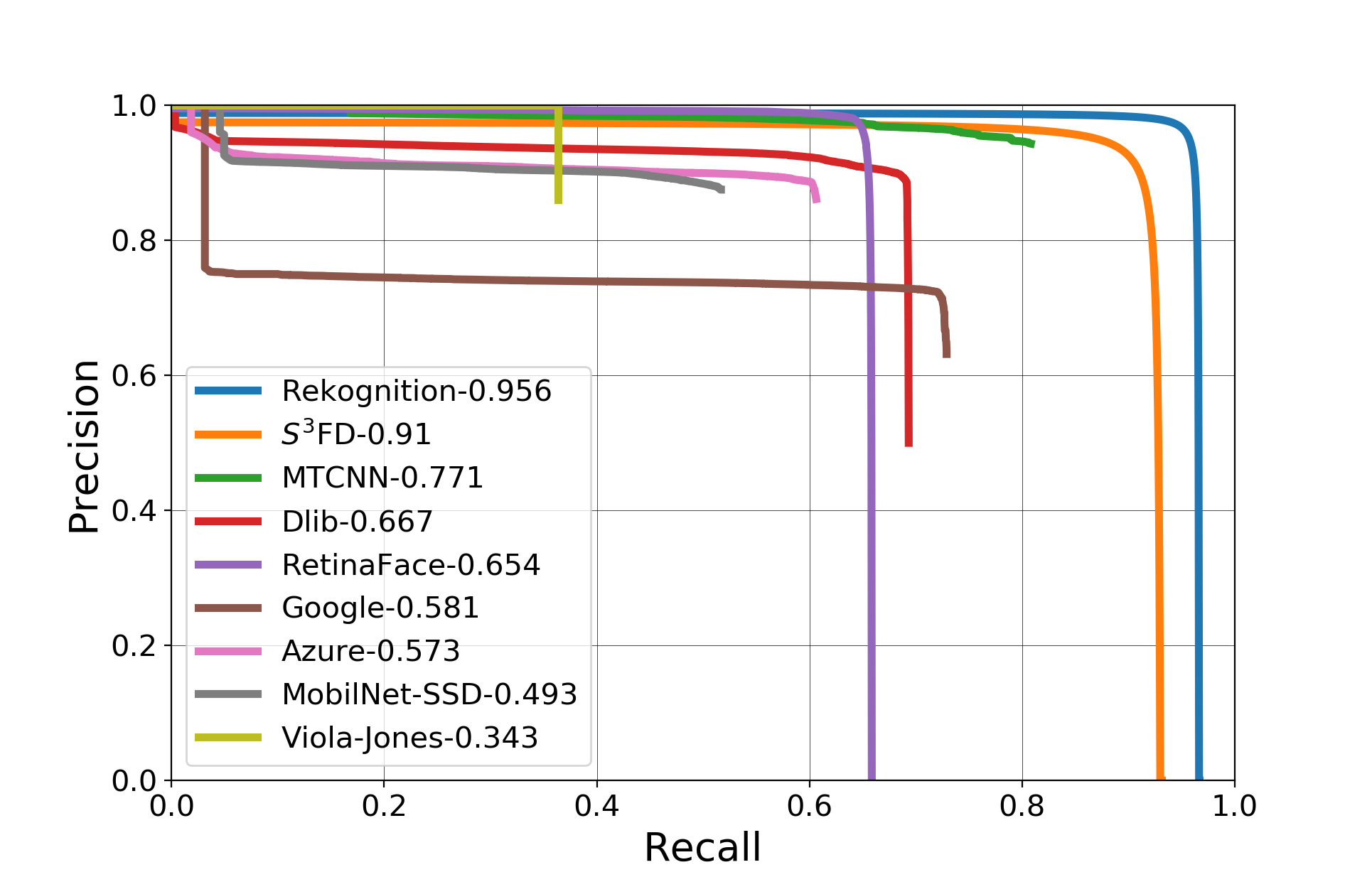}
    \caption{Precision-recall curves of face detection algorithms with accuracy.}
\end{figure}

\subsection{Five Facial Landmark Accuracy}

Concisely, the purpose of facial landmark extraction is to detect facial landmark coordinates in a given face, inside of the bounding box. To evaluate the accuracy of five facial landmark localisation, we compare Amazon Rekognition and MTCNN on the DHA. MTCNN is widely used in many facial analysis tasks, such as face verification \cite{wang2018additive} and face recognition \cite{zhang2014facial, wang2018cosface}. Amazon Rekognition software returns thirty facial landmarks, however, we consider only five facial landmarks as used in MTCNN to compare their performance. We excluded other face-detection algorithms due to their poor performance on detecting faces in the DHA dataset. Here, we employ the face box size ($\sqrt{\emph{W} \times \emph{H}}$) as the normalisation distance. As shown in Fig. 7, we give the mean error of each facial landmark on the DHA dataset. MTCNN achieves significantly lower mean error (NME) than Amazon Rekognition, respectively 6.39\% and 15.94\%. However, the detection rate of MTCNN is 77\% which is significantly lower than Amazon Rekognition with 95\%. There is an extensive trade off between detecting faces and localisation of landmarks. These results show that both face detection algorithms fail to provide sufficient performance on both detecting faces and localising five facial landmarks. 
\begin{figure}[t]
\centering
    \includegraphics[width=8.3cm, height=7cm]{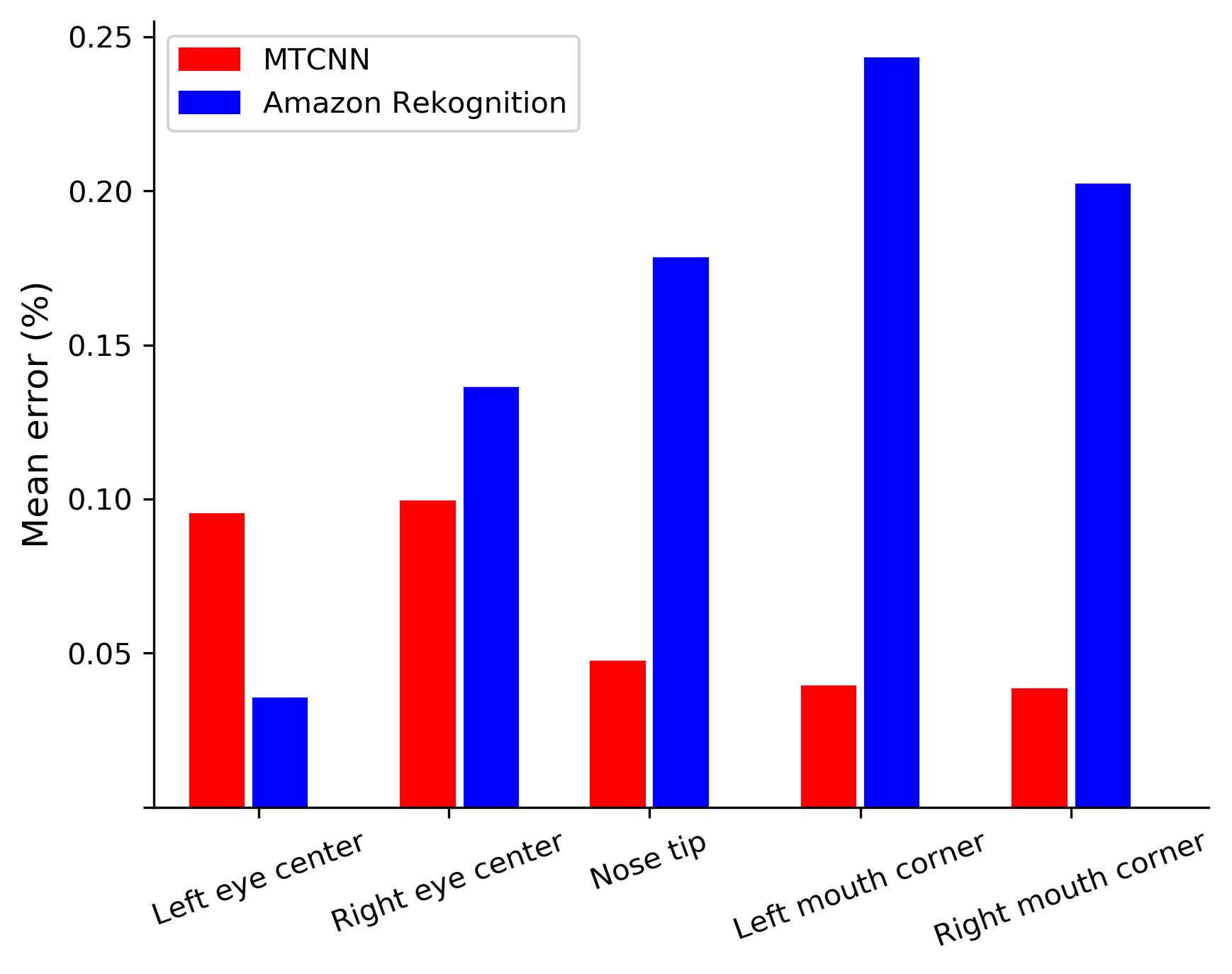}
    \caption{Comparison of five facial landmark localisation between Amazon Rekognition and MTCNN on the DHA.}
\end{figure}
\section{Conclusion}

Labeled face image datasets are foundational for developing automated facial analysis systems that play a major role in our lives. Constructing a labeled face image dataset is often difficult, costly, and current methods of image collection can be biased towards over-represented individuals. However, individuals differ in many ways, and it is hard to predict how the system will perform on faces underrepresented in the training set. The importance of developing fair systems is becoming more important as these systems play a larger role in our society. In this work, we introduced the Distinctive Human Appearances dataset to evaluate face  algorithms and demonstrate the importance of capturing lower-frequency human appearances. Compared to other face image datasets, DHA contains only one face per image, has high-quality, and no occlusion, which are normally favorable conditions for face detection, enabling evaluation of face-detection performance with respect to diverse human appearances alone, without other degrees of variability.
 
We evaluated human annotators' performance and the representatives of popular face detection algorithms: Viola-Jones, Dlib as traditional methods, MTCNN, MobilNet-SSD as deep learning models, Amazon Rekognition, Microsoft Azure, and Google Cloud Vision as cloud-based face detection software. Compared to previously reported performance evaluations, these algorithms performed much worse on the DHA dataset. These results raise a concern about the real-world impact of these algorithms. In addition, our results show that human performance on face annotation can be worse in cases of extreme body modification and when  faces are partially covered. Because these systems are trained on human annotations, the quality of the annotations becomes essential, particularly when the annotation quality is worse for more distinctive appearances. 

Fairness and bias are an important aspect to consider in designing and developing computer vision systems, especially because these systems play an important role in our lives. Although large-scale image datasets have contributed to the recent improvements in accuracy, there are concerns about their fairness and transparency. The diverse dataset proposed in this paper will help reveal discover gaps in the capabilities of face detection technologies and mitigate the bias existing in widely used computer vision systems.



\normalsize

\printbibliography{}
\end{document}